\documentclass[11pt,a4paper]{article}
\usepackage[hyperref]{emnlp2020}
\usepackage{times}
\usepackage{latexsym}

\usepackage[utf8x]{inputenc}
\usepackage[thaicjk,english]{babel}
\addto\extrasthaicjk{\fontencoding{C90}\selectfont}
\makeatletter
\@namedef{opt@inputenc.sty}{utf8}
\makeatother
\usepackage{CJKutf8}
\usepackage{amsmath}
\usepackage{amsthm}
\usepackage{amsfonts}
\usepackage{amssymb}
\usepackage{mathtools}

\usepackage{graphicx}
\usepackage{booktabs}
\usepackage{tabularx}
\usepackage{tabulary}
\usepackage{colortbl}
\usepackage{xspace}
\usepackage{xcolor}
\usepackage{multirow}
\usepackage{comment}
\usepackage{microtype}
\usepackage{arydshln}
\usepackage{tikz}
\def\checkmark{\tikz\fill[scale=0.4](0,.35) -- (.25,0) -- (1,.7) -- (.25,.15) -- cycle;} 

\newcolumntype{Y}{>{\centering\arraybackslash}X}

\usepackage{url}
\usepackage{enumitem}
\fboxsep=-\fboxrule
\usepackage{scalerel}
\def\phraseA{\scalerel*{%
  \setbox0=\hbox{\raisebox{-9pt}{\includegraphics{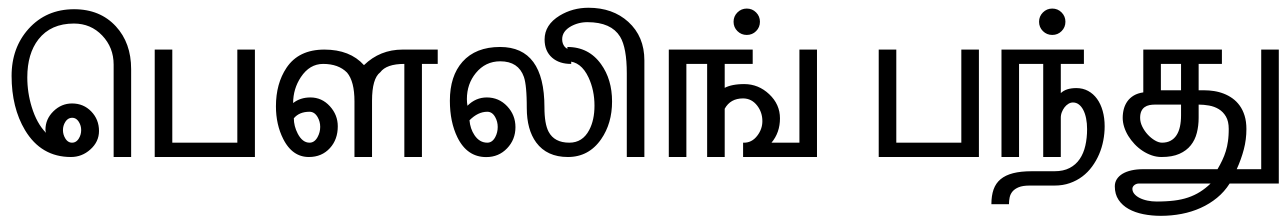}}}\dp0=0pt\box0}{X}}
\def\phraseB{\scalerel*{%
  \setbox0=\hbox{\raisebox{-2pt}{\includegraphics{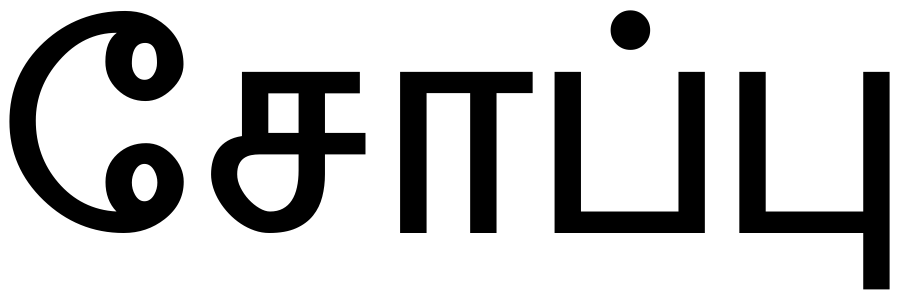}}}\dp0=0pt\box0}{X}}
  \def\phraseC{\scalerel*{%
  \setbox0=\hbox{\raisebox{-9pt}{\includegraphics{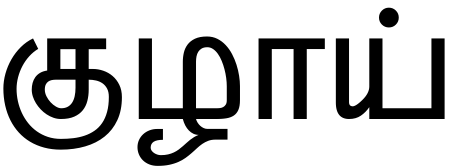}}}\dp0=0pt\box0}{X}}

\aclfinalcopy % Uncomment this line for the final submission

\newcommand{\ignore}[1]{}

\newcommand{\alert}[1]{{\color{red}[TODO]#1}}

\title{XCOPA: A Multilingual Dataset for Causal Commonsense Reasoning}

\author{Edoardo M. Ponti$^1$\thanks{~~Equal contribution.},  Goran Glava\v{s}$^{2}$\footnotemark[1], ~Olga Majewska$^1$, \\ {\bf Qianchu Liu$^1$, ~Ivan Vuli\'{c}$^{1}$,  ~Anna Korhonen$^1$} \smallskip \\
$^1$Language Technology Lab, TAL, University of Cambridge, UK \\
$^2$ Data and Web Science Group, University of Mannheim, Germany \\
$^1$\texttt {\{ep490,om304,ql261,iv250,alk23\}@cam.ac.uk} \\
$^2$\texttt {goran@informatik.uni-mannheim.de}
}
\date{}

\begin{document}
\maketitle

\begin{abstract}

% Lack of datasets for commonsense reasoning in the form of multiple-choice classification. Lack of multilingual datasets.
% Baselines show that current state-of-the-art models lack ability to cross-lingual transfer this faculty: advance hypotheses `curse of multilinguality'. 
%Human communication is rooted in the speakers’ knowledge and understanding of the world, and much of the exchanged information is implicit in the transmitted signal. 
In order to simulate human language capacity, natural language processing systems must be able to reason about the dynamics of everyday situations, including their possible causes and effects. Moreover, they should be able to generalise the acquired world knowledge to new languages, modulo cultural differences. Advances in machine reasoning and cross-lingual transfer depend on the availability of challenging evaluation benchmarks. Motivated by both demands, we introduce Cross-lingual Choice of Plausible Alternatives (\textbf{XCOPA}), a typologically diverse multilingual dataset for causal commonsense reasoning in 11 languages, which includes resource-poor languages like Eastern Apurímac Quechua and Haitian Creole. We evaluate a range of state-of-the-art models on this novel dataset, revealing that the performance of current methods based on multilingual pretraining and zero-shot fine-tuning falls short compared to translation-based transfer. Finally, we propose strategies to adapt multilingual models to out-of-sample resource-lean languages where only a small corpus or a bilingual dictionary is available, and report substantial improvements over the random baseline. The XCOPA dataset is freely available at \href{https://github.com/cambridgeltl/xcopa}{\texttt{github.com/cambridgeltl/xcopa}}.
\end{abstract}

\section{Introduction}
\label{s:introduction}
\begin{table*}[!t]
    \centering
    {\small
    \begin{tabularx}{1.0\textwidth}{cYcYY}
\toprule
\textsc{} & \textsc{premise} & \textsc{} & \textsc{choice 1} & \textsc{choice 2} \\
\hline
qu & \textit{Sipasqa cereal mikhunanpi kuruta tarirqan.} & \multirow{3}{*}{R} & \textit{Payqa pukunman ñuqñuta churakurqan.} & \textit{Payqa manam mikhuyta munarqanchu.} \\
en & {The girl found a bug in her cereal.} & & {She poured milk in the bowl.} & {She lost her appetite.} \\
\hline
th & \foreignlanguage{thaicjk}{ตาของฉันแดงและบวม} &\multirow{2}{*}{C} & \foreignlanguage{thaicjk}{ฉันร้องไห้} & \foreignlanguage{thaicjk}{ฉันหัวเราะ} \\
en & {My eyes became red and puffy.} & & {I was sobbing.} & {I was laughing.} \\
\bottomrule
    \end{tabularx}
    }%
    \vspace{-1.5mm}
    \caption{Examples of forward (Result [R]) and backward (Cause [C]) reasoning from the XCOPA validation sets.}
    \label{tab:causeeffect}
    \vspace{-1.5mm}
\end{table*}

% Explosion of commonsense datasets, nothing beyond English
%% based on deductions from the evidence presented in texts
Commonsense reasoning is a critical component of any natural language understanding system \cite{Davis:2015cacm}. Contrary to textual entailment, commonsense reasoning must bridge between premises and possible hypotheses with \textit{world knowledge} that is not explicit in text \citep{singer1992validation}. Such world knowledge encompasses, among other aspects: temporal and spatial relations, causality, laws of nature, social conventions, politeness, emotional responses, and multiple modalities. Ultimately, it shapes the individuals' expectations about typical situations \citep{shoham1990nonmonotonic}.\footnote{Moreover, there are often multiple legitimate chains of sentences that can be invoked in between premises and hypotheses. In short, commonsense reasoning does not just involve understanding what is possible, but also ranking what is most \textit{plausible}.}

A seminal work on the quantitative evaluation of commonsense reasoning is the Choice Of Plausible Alternatives dataset \citep[COPA;][]{Roemmele:2011aaai}, which focuses on cause--effect relationships. In recent years, more datasets have been dedicated to other facets of world knowledge \citep[\textit{inter alia}]{Sakaguchi:2020aaai,Bisk:2020aaai,Bhagavatula:2020iclr,Rashkin:2018acl,Sap:2019siqa}. Unfortunately, the extensive efforts related to this thread of research have so far been limited only to the English language.\footnote{The only exception is direct translation of the 272 paired English Winograd Schema Challenge instances to Japanese \cite{WSCja}, French \cite{Amsili:2017french}, and Portuguese \cite{Melo:2020eniac}.} Such a narrow scope not only curbs the development of natural language understanding tools in other languages \cite{Bender:2011lilt,Ponti:2019cl}, but also exacerbates the Anglo-centric bias in modeling commonsense reasoning. In fact, the expectations about typical situations do vary across cultures \citep{thomas1983cross}.

% Cross-lingual datasets for NLU.
Datasets that cover multiple languages for other natural understanding tasks, such as language inference \citep{Conneau:2018emnlp}, question answering \citep{Lewis:2019arxiv,artetxe2020translation,tydiqa}, and paraphrase identification \citep{Yang:2019emnlp} have received increasing attention. In fact, the requirement to generalise to new languages encourages the development of more versatile language understanding models, which can be ported across different grammars and lexica. These efforts have recently culminated in the integration of several multilingual tasks into the XTREME evaluation suite \cite{Hu:2020arxiv}. However, 
%despite the abundant and steadily growing list of multilingual evaluation sets, 
a comprehensive multilingual benchmark for commonsense reasoning in particular is still missing. 
%XCOPA aims to fill this gap, respecting typological diversity by covering a representative language set, and also including some truly low-resource languages.

% We fill the gap. Purpose of language sample: generalising to resource-rich languages vs generalising to any language. Sampling speakers vs sampling languages.
In order to address this gap, we develop a novel dataset, %that connects cross-lingual transfer and causal commonsense reasoning: 
\textbf{XCOPA} (see examples in Table~\ref{tab:causeeffect}), by carefully translating and re-annotating the validation and test sets of English COPA into 11 target languages. A key design choice is the selection of a typologically diverse sample of languages. In particular, we privilege variety over the abundance in \textit{digital} texts. 
%In fact, the samples in most multilingual evaluation benchmarks reflect the abundance in \textit{digital} resources of languages, i.e., their diffusion on the internet, rather than representing the actual cross-lingual variation in the world's languages. 
Since resource-rich languages tend to belong to a few families and areas, samples inspired by this criterion are highly biased and not indicative of true models' performance \citep{gerz-etal-2018-relation,Ponti:2019cl,Joshi:2020arxiv,lauscher2020zero}.
Following this guiding principle, we select 11 languages from 11 distinct families, and 5 geographical macro-areas (Africa, Eurasia, Papunesia, North America, and South America). 

%% (IV, removed, we'll mention how the dataset looks like later on)
%%As shown in Table~\ref{tab:causeeffect}, each instance corresponds to a premise, a prompt (``\textit{What was the \textsc{cause}}?'' or ``\textit{What happened as a \textsc{result}}?''), and two choices. The task consists in a binary classification where the machine has to predict the more plausible choice.

%Language sampling 

% Benchmarks and extension of pre-trained models.
We leverage XCOPA to benchmark a series of state-of-the-art pretrained multilingual models, including \textsc{xlm-r} \citep{conneau2019unsupervised}, \textsc{mbert} \citep{Devlin:2019bert}, and multilingual \textsc{use} \citep{yang2019multilingual}. Two XCOPA languages (i.e., Southern Quechua and Haitian Creole) are out-of-sample for the pretrained models: this naturally raises the question of how to adapt the pretrained models to such unseen languages. 
%, and the catastrophic forgetting effect on old languages. 
In particular, we investigate the resource-lean scenarios where either some monolingual data or a bilingual dictionary with English (or both) are available for the target language. 

In summary, we offer the following contributions. \textbf{1)} We create the first large-scale multilingual evaluation set for commonsense reasoning, spanning 11 languages, and discuss the challenges in accounting for world knowledge across different cultures and languages. \textbf{2)}  We propose quantitative metrics to measure the internal variety of a language sample, which can guide the design of any multilingual dataset in the future. \textbf{3)} We benchmark pretrained state-of-the-art models in cross-lingual transfer of commonsense knowledge, and \textbf{4)} investigate how to (post-hoc) improve transfer for languages unseen at pretraining time. 

%\begin{itemize}
%    \item 
%    \item
%    \item 
%\end{itemize}

In order to rise to the challenge of this dataset, models must be able not only to combine textual evidence with world knowledge -- which makes commonsense reasoning challenging \textit{per se} \cite{Talmor:2019naacl,Rajani2019acl}, but they must also transfer the acquired causal reasoning abilities across languages. The results we obtain on XCOPA thus indicate the limitations of current state-of-the-art multilingual models in cross-lingual transfer settings for complex reasoning tasks. 
%the progress of the NLP field on achieving these key goals.

%IVAN: This should be mentioned in the introduction, and also aligns well with the story I'm telling in the RW section -- 1. There are plenty of commonsense reasoning datasets, mushrooming like crazy in the last 1-2 years, but there is absolutely nothing beyond the English language; 2. There are so many cross-lingual evaluation datasets, also with much increased interest in last years, but there is nothing multilingual and comprehensive for commonsense. XCOPA addresses these important gaps. Work on commonsense reasoning in multilingual settings is still undeveloped: current research is still focused predominantly on English. We connect cross-lingual NLP and commonsense reasoning.

\section{Annotation Design}
\label{s:annotation}
%% behind our chosen dataset creation approach 
%% with cross-lingual alignments (in order 
\noindent \textbf{Design Objectives.}
The principal objectives in devising XCOPA were: \textbf{1)} to create a \emph{genuinely} typologically diverse multilingual dataset, aligned across target languages to make performance scores comparable, and \textbf{2)} to ensure high quality, naturalness and idiomacity of each monolingual dataset. While the commonly used translation approach addresses the former objective, 
%i.e., it ensures mappings between the source and target language, 
it is prone to compromise the latter goal, bending the target language to the structural and lexical properties of the source language: the resulting evaluation benchmarks thus fail to 
%capture the characteristics of the target language accurately and 
measure system performance adequately \cite{koppel2011translationese,volansky2015features,artetxe2020translation,freitag2020bleu}.

To avoid these pitfalls, we: 
%and ensure both objectives are met were to 
(i) entrusted the translation task to a single (carefully selected) translator per target language,\footnote{Crowd-sourcing offers faster annotation at a lower cost -- however, in our trial experiments, chasing low annotation times and costs resulted in low translation quality. In our experiment, the average wage was £15 per hour, for a (self-paced) total time per language between 12 and 20 hours.} and (ii) offered enough leeway for necessary target-language adjustments (e.g., substitutions with culture-specific concepts and multi-word paraphrases, wherever the original text eluded direct translation).\ignore{Allowing for language-specific modifications ensured that the output is natural and understandable to native speakers.} Detailed translation guidelines are available in the Appendix.%~\ref{ss:dataguide}. %%in the target language and that mentioned concepts are understandable to native speakers of that language.
%%
%%% This can be shortened
%% (IV, removed now)
%% Applying more stringent eligibility criteria (see \S \ref{ss:dataguide} for details) and maintaining direct communication with the translators throughout the process facilitated quality control and allowed us to resolve problems on a case-by-case basis (see \S \ref{s:qualitative} for detailed discussion). Allowing for language-specific modifications ensured that the output is natural in the target language and that mentioned concepts  are understandable to native speakers of that language.

%- while preserving the original causal relations and cross-lingual sentence-level substitutability of the dataset instances.

\begin{table*}[ht]
    \def\arraystretch{0.91}
    \centering
    {\small
    \begin{tabularx}{1.0\textwidth}{lY YYYYYY}
    \toprule
{} & Range & XCOPA & TyDiQA & XNLI & XQUAD & MLQA & PAWS-X \\
    \cmidrule(lr){3-8}
Typology & [0, 1] & {0.41} & {0.41} & 0.39 & 0.36 & 0.32 & 0.31 \\
Family & [0, 1] & {1} & 0.9 & 0.5  & 0.6 & 0.66 & 0.66 \\
Geography & [0, $\ln 6$] &  {1.67} & 0.92 & 0.37 & 0 & 0 & 0 \\

\bottomrule
    \end{tabularx}
    }%
    \vspace{-1.5mm}
    \caption{Indices of typological, genealogical, and areal diversity for the language samples of a set of NLU datasets.}
    \label{tab:avgentropies}
    \vspace{-1.5mm}
\end{table*}

\vspace{1.4mm}
\noindent \textbf{Language Sampling.}
\label{ss:languagesampling}
Multilingual evaluation benchmarks assess the \textit{expected performance} of a model across languages. However, should such languages be sampled according to the distribution of digital texts or rather based on the distribution over the languages spoken around the world? The former strategy is unreliable, as languages rich in resources tend to belong to the same families and areas, which facilitates knowledge transfer and hence leads to an overestimation of the expected performance \citep{gerz-etal-2018-relation,Ponti:2019cl}. %Moreover, 

Moreover, rather than samples that account for independent and identically distributed draws from the `true' language distribution (known as \textit{probability} sampling), we opt for a \textit{uniform} distribution of linguistic phenomena, which encourages the inclusion of outliers \citep[known as \textit{variety} sampling;][]{rijkhoff1993method,dryer1989large}. Thus, the performance on XCOPA also reflects the \textit{robustness} of a model, i.e., its resilience to linguistic features that are unlikely to be observed in the training data.

%The latter approach has the advantage of being unbiased. In fact, 

%% different assumptions about the desired distribution over languages lie under the choice of the language sample. 
%% If so, resource-rich languages should be represented more conspicuously.

%probability samples variety samples. Representative (simulates independently drawn) vs inclusivity of rare phenomena. Expected performance vs robustness. Survey by \citet{}. 

%% lthough several sophisticated methods have been developed for diversity sample, such as Diversity Value \citep{rijkhoff1993method} and Genus-Macroarea \citep{miestamo2004clausal}

Inspired by \citet{rijkhoff1993method} and \citet{miestamo2004clausal}, we propose a series of simple and interpretable metrics that quantify diversity of a language sample independent of its size: \textbf{1)} a \textit{typology} index based on 103 typological features of each language from URIEL \cite{Littel-et-al:2017}, originally sourced from the World Atlas of Language Structures \citep[WALS;][]{wals}. Each feature is binary and indicates the presence or absence of an attribute in a language. We estimate the entropy of the distribution of values in a sample. %, as shown in the heatmap of Figure~\ref{fig:entropiesbyfeat} in the Appendix.
Afterwards, we average across all 103 feature-specific entropies. Intuitively, if all values are equally represented, the entropy is high. If all languages have identical features, the entropy is 0; \textbf{2)} The \textit{family} index is simply the number of distinct families divided by the sample size. \textbf{3)} The \textit{geography} index is the entropy of the distribution over macro-areas in a sample.\footnote{Six macro-areas, as defined by WALS \cite{wals}, are: Africa, Australia, Eurasia, North America, Papunesia, and South America. Whenever a language spans multiple macro-areas, we select that of the standard variety.}

\begin{table*}[!t]
    \def\arraystretch{0.94}
    \centering
    {\small
    \begin{tabularx}{1.0\textwidth}{l|YYYYYYYYYYY}
    \toprule
& \textsc{et} & \textsc{ht} & \textsc{id} & \textsc{it} & \textsc{qu} & \textsc{sw} & \textsc{ta} & \textsc{th} & \textsc{tr} & \textsc{vi} & \textsc{zh}\\
\cmidrule(lr){2-12}
\textit{val} & 97.0 & 97.0 & 99.0 & 98.0 & 98.0 & 99.0 & 100.0 & 99.0 & 97.0 & 97.0 & 96.0\\
\textit{test} &  98.2 & 96.4 & 100.0 & 97.0 & 94.8 & 99.0 & 98.6 & 98.2 & 96.4 & 98.4 & 96.6\\
\bottomrule
    \end{tabularx}}%
    \caption{Percentage of annotated labels in each language agreeing with the majority label. Note that the majority label is highly reliable, as we observed a 100\% agreement with the development set labels in the original COPA.}
    \label{tab:overlap}
    \vspace{-1.5mm}
\end{table*}

The sample of languages for XCOPA aims at maximising these indices. In particular, XCOPA includes Estonian (\textsc{et}), Haitian Creole (\textsc{ht}), Indonesian (\textsc{id}), Italian (\textsc{it}), Eastern Apurímac Quechua (\textsc{qu}), Kiswahili (\textsc{sw}), Tamil (\textsc{ta}), Thai (\textsc{th}), Turkish (\textsc{tr}), Vietnamese (\textsc{vi}), and Mandarin Chinese (\textsc{zh}). These languages belong to distinct families, respectively: Uralic, Creole, Austronesian, Indo-European, Yuman–Cochimí, Niger-Congo, Dravidian, Kra-Dai, Turkic, Austroasiatic, and Sino-Tibetan. Moreover, \textsc{ht} and \textsc{qu} are spoken in North and South America, respectively, which are both underrepresented macro-areas. We report the 3 metrics in Table~\ref{tab:avgentropies} and compare them to samples from other standard multilingual NLU datasets. XCOPA offers the most diverse sample in terms of typology (on a par with TyDiQA), family, and geography. %also shows very high scores.

\vspace{1.4mm}
\noindent \textbf{Final Dataset.}
As Table~\ref{tab:causeeffect} shows, each (X)COPA instance corresponds to a premise, a question (``\textit{What was the \textsc{cause}}?'' or ``\textit{What happened as a \textsc{result}}?''), and two alternatives. The task is framed as binary classification where the machine has to predict the more plausible choice. For each target language, XCOPA comprises 100 annotated data instances in the validation set and 500 instances as the test set, which are translations from the respective English COPA validation and test set. Our translators performed labeling prior to translation, deciding on the correct alternative for the English premise and preserving the correctness of the same alternative in translation. We measure inter-translator agreement using the Fleiss' $\kappa$ statistic \cite{fleiss1971measuring}: the obtained score of $0.921$ for validation and $0.911$ for test data reveal very high agreement (i.e., \citet{landis1977measurement} define $\kappa\geq0.81$ as almost perfect agreement).

From the 11 sets of annotation labels we obtain the majority labels (i.e., 6+ translators agree). We observe perfect agreement between our majority labels and the English COPA labels for development data. We then compute the percentage of annotated labels which agree with the majority label for each language individually, reported in Table \ref{tab:overlap}, and find very high agreement across 11 languages. The small discrepancies in label choices in our work stem not only from the actual semantic ambiguity of the original English question, but also reflect the translators' different cultural frames of reference and patterns of association. On average, 2.1\% of labels in the validation set and 2.4\% of labels in the test set do not match the majority label.\footnote{In order to accurately represent ambiguity of the small number of disagreement labels, in the final datasets we explicitly tag the corresponding questions with an apposite marker.} 

\section{Qualitative Analysis}
\label{s:qualitative}
%\textcolor{red}{Edo / Olga}

%Differences in performance across different languages.

%\subsection{Translation challenges}
As highlighted in \S\ref{s:annotation}, our guidelines anticipated that the adopted translation approach may entail language-specific challenges, e.g., the lack of equivalent concepts or the grammatical expression of tense and aspect. We now analyse the main design challenges and the adopted solutions.
% Are single words or rather idiomatic expressions causing translation problems?
% Are there particular semantic domains which cause difficulties in particular languages, and are there some core concepts which are universally shared and understood? (e.g., legal terminology versus emotions, body parts)
%% (IV, removed)
%% Another source of translation difficulties lay in discrepancies in grammatical features such as tense or aspect (e.g., lack of verbal tense in the target language or differences and regional variation in usage of particular tensed forms).

%\paragraph{Different cultural context}
\vspace{1.4mm}
\noindent \textbf{Cultural Context.}
The scenarios included in English COPA were authored by American English speakers with a particular cultural background. It is therefore inevitable that some concepts, intended as commonplace, sound unusual or even completely foreign in the target language. Examples include: (i) concrete referents with no language-specific term available (e.g., \textit{bowling ball}, \textit{hamburger}, \textit{lottery}); (ii) systems of social norms absent in the target culture, e.g., traffic regulations (e.g., \textit{parallel parking}, \textit{parking meter}); (iii) social, political, and cultural institutions and related terminology (e.g., e.g.,
\textit{mortgage}, \textit{lobbyist}, \textit{gavel}); (iv) idiomatic expressions (e.g., \textit{put the caller \underline{on hold}}). %It is therefore inevitable that some of the concepts featured in the dataset, intended as commonplace, will be unusual or even completely foreign in the target language and culture. Examples flagged by our translators include:

In such cases, the translators were advised to resort (in the following order of preference) to (i) paraphrasing; (ii) substitutions with similar concepts, e.g., `faucet' is replaced with `pipe' in Tamil (\phraseC{}, \textit{kuḻāy}) and Haitian Creole (\textit{tiyo}); or (iii) phonetically transcribed loan words, e.g., in Tamil: \phraseA{} (\textit{\underline{pauliṅ} pan\-tu}, `\underline{bowling} ball'), \phraseB{} (\textit{cōppu}, `soap'). 

\vspace{1.4mm}
\noindent \textbf{Grammatical Tense.} 
%% (IV, this is interesting, but too verbose)
%%The scenarios included in COPA refer to events that took place in the past and are formulated in what can be described as a narrative register (one of the sources from which question topics were drawn was a corpus of personal stories published online \cite{gordon2009identifying}). This is grammatically rendered exclusively by means of past simple (preterite) or past continuous (imperfect) verb forms. Temporal anteriority of a hypothesis sentence with respect to the premise is not grammatically marked (e.g., with a past perfect verb form) and can only be deduced based on the prompt (``\textit{What was the \textsc{cause} of this?}''). The preterite-imperfect contrast used in English to distinguish background states (imperfective) from the main event (perfective) (e.g., \textit{I was expecting company.} \textsc{imperf} vs. \textit{I tidied up my house.} \textsc{perf}) is not universally applicable and different languages employ different discourse grounding strategies \cite{hopper1979aspect}, which has interesting implications for the multilingual extension of the dataset.
%% (e.g., in Chinese, temporally bounded situations are typically interpreted as having past tense unless otherwise indicated \cite{smith2000presented}).
%% , where they were the translator's preferred choice over the marked counterparts
The temporal contiguity between two events and their duration is crucial in establishing their causal relationship \citep{enfieldmacro}. A number of languages in our sample (i.e., \textsc{th}, \textsc{vi}, \textsc{id}, \textsc{zh}) do not have the grammatical category of tense and express temporality by means of aspect, mood or lexical items and expressions referring to time (e.g., adverbs), or rely entirely on pragmatic context to provide sufficient information for the interpretation of the utterance. Even if aspectual viewpoint markers exist, they are optional, e.g., the perfective marker \begin{CJK}{UTF8}{min}了\end{CJK} (\textit{le}) in \textsc{zh}. To ensure naturalness of the translated sentences and faithfully represent the properties of the so-called tenseless languages, we favoured the unmarked variants, with the temporal relations established by the situational context. For example, compare: (a) \begin{CJK}{UTF8}{gbsn}我想节约能源。\end{CJK}, \textit{Wǒ xiǎng jiéyuē néngyuán.}, `I want(ed) to conserve energy.' (no perfective marker), and (b) \begin{CJK}{UTF8}{gbsn}学生拼错了这个词。\end{CJK}, \textit{Xuéshēng pīn cuò\textbf{le} zhège cí.}, `The student misspelled the word.' (with completed action marker). %Further considerations on anteriority and aspect are provided in Appendix~\ref{ss:gramtense}.

\vspace{1.4mm}
\noindent \textbf{Label Discrepancies.}
The analysis of inter-translator agreement in \S\ref{s:annotation} revealed a small number of COPA scenarios with discrepancies in annotations across languages. To better understand the source of such disagreements, we identified all the validation set instances on which one or more translators diverged from the majority label.\footnote{Overall, there were 10 validation set questions with 1 translator out of 11 in disagreement, 5 questions with 2, and 1 question with 3. %The latter is a scenario characterised by a high degree of ambiguity regardless of cultural context.
%(i.e., an anomaly in the case could either result in the detective finalizing or scrapping his theory), where the Estonian, Swahili, and Turkish translators settled for Choice 1 rather than Choice 2.
} We identified two cases where the translator's experience and cultural frame of reference played a role (as attested in translator feedback), which required, for instance, understanding of the procedures and structure of U.S. court trials (e.g., \textit{The judge pounded the gavel.} \textsc{cause}: (a) \textit{The courtroom broke into uproar.} (b) \textit{The jury announced its verdict.}). %or familiarity with parking enforcement regulations and customs typical in the U.S. but absent in the target culture (\#92 in Table~\ref{tab:maxdisagree}). %(e.g., \textit{The man received a parking ticket.}, \textsc{cause}: (a) \textit{He parallel parked on the street.}; (b) \textit{The parking meter expired.}). 

Most disagreement cases (87.5\%), however, seem to be culturally independent and concern genuinely ambiguous cases (e.g.\ \textit{The detective revealed an anomaly in the case.} \textsc{result}: (a) \textit{He finalized his theory.} (b) \textit{He scrapped his theory.}). To verify this in a monolingual setting, we conducted a follow-up experiment where 4 Italian native speakers labeled the translated validation and test instances. The Fleiss' $\kappa$ agreement scores were $0.926$ (validation) and $0.917$ (test).
This corroborates 
%our observation about a certain number of sentences being ambiguous regardless of cultural considerations
%This analysis shows that the original ambiguity carries over to the translated sentences and both alternatives are plausible candidates. 
%It supports 
our decision to override a single translator's label with the majority label without altering the translation. % label's validity for a given outlier language.

%% and alternative pairs with near equal degree of plausibility

\section{Experiments and Results}
\label{s:experiments}
XCOPA is a multiple--choice classification task: given a premise and a prompt (\textsc{cause} or \textsc{result}), the goal is to select the more plausible of the two answer choices (see Table~\ref{tab:causeeffect}). We now benchmark a series of state-of-the-art models on XCOPA to provide baseline scores for future research, as well as to expose the challenging nature of the dataset.  In \S\ref{ssec:mcclas}, we list the main axes of comparison of our baselines and outline the general neural architecture for multiple-choice classification that we employ in all experiments. In \S\ref{ssec:resdisc}, we discuss the results. Afterwards in \S\ref{ssec:advvar}, in order to prove that solving this task requires relying on true causal reasoning rather than spurious correlations, we test the best-performing model on `adversarial' variants where either the premise or the prompt are hidden \citep{niven-kao-2019-probing}. Finally, in \S\ref{s:extension} we explore several strategies to adapt massively multilingual models to new languages not observed during pretraining, such as Quechua and Haitian Creole.

%% (IV, we don't need this)
\iffalse
Pretrained neural language encoders come in two main flavors: (a) in the feature-based paradigm, the encoder provides the fixed representation of an input text, which is then fed to the task-specific classifier: the parameters of the encoder do not change in downstream training \cite{peters2018deep,yang2019multilingual}; (b) in the more popular fine-tuning approach, the parameters of the encoder are fine-tuned together with the classifier parameters in the downstream training \cite{Devlin:2019bert,liu2019roberta,conneau2019unsupervised,clark2019electra}. For monolingual English downstream tasks with (reasonably) large training sets, fine-tuning has been shown to perform better than feature-based encoding. 
%%
In our experiments on XCOPA (i.e., cross-lingual transfer for causal commonsense reasoning), instead, we require multilingual encoders and have only (1) a small English COPA training set consisting of only 400 instances and (2) an even smaller development set in each target language (encompassing mere 100 instances). 
\fi

\subsection{Baselines}
\label{ssec:mcclas}
We evaluate baselines in several combinations of experimental setups based on: \textbf{1)} different methods for cross-lingual transfer, either based on model transfer or machine translation; \textbf{2)} different multilingual pretrained encoders; \textbf{3)} different sources of training and validation data.

\vspace{1.4mm}
\noindent \textbf{Cross-lingual Transfer Methods.}
We consider two high-level methods for cross-lingual transfer \citep{tiedemann-2015-cross,Ponti:2019cl}: \textbf{1)} \textit{multilingual model transfer} (MuMoTr), whereby a Transformer-based encoder is pretrained on multiple languages in an unsupervised fashion, and subsequently trained on English annotated data for multiple-choice classification, therefore enabling zero-shot generalisation to the other languages. \textbf{2)} \textit{translate test} (TrTe), whereby target test data\footnote{As shown by \citet{Conneau:2018emnlp}, translating the test data from the target language to English is more cost-effective than translating English training data into the target language.} are translated into English via Google Translate. This includes all languages except for \textsc{qu}, for which the service is not available.

%\footnote{Via the API \url{pypi.org/project/googletrans/}.} %In this case, we resort to a monolingual English pretrained encoder.

\vspace{1.4mm}
\noindent \textbf{Multilingual Encoders.}
For model transfer, we evaluate the following state-of-the-art pretrained multilingual encoders: \textbf{1)} multilingual BERT (MBERT) \cite{Devlin:2019bert} and XLM-on-RoBERTa \cite{conneau2019unsupervised}, both the Base (XLM-R) and Large (XLM-R-L) variants, in the standard \textit{fine-tuning regime} (i.e., their parameters are fine-tuned together with the task classifier's parameters), and \textbf{2)} multilingual Universal Sentence Encoder (USE) \cite{yang2019multilingual} in the \textit{feature-based regime} (i.e., its parameters are fixed during the task classifier's training). Both MBERT and XLM-R include all XCOPA languages in their pretraining data spanning $\sim$100 languages, except Haitian Creole and Quechua. Multilingual USE was trained on 16 languages, covering \textsc{it}, \textsc{th}, \textsc{tr}, and \textsc{zh} from the XCOPA language sample.

\vspace{1.4mm}
\noindent \textbf{Data Sources.} The only direct in-domain data available for training is the original English COPA training set covering 400 instances. 
Due to data scarcity, we probe the usefulness of an intermediate training stage \citep{phang2018sentence,glavavs2020supervised} on larger multiple--choice English commonsense reasoning datasets, such as \textsc{SocialIQa} \cite[SIQA;][]{Sap:2019siqa}, before fine-tuning the model on COPA. The SIQA dataset is in a distant domain (commonsense reasoning about social interactions) with open-format prompts and three answer choices. However, it comes with a much larger training set, consisting of 33K instances. Therefore, it can provide useful learning signal also for causal commonsense reasoning in XCOPA. Moreover, we consider two different model selection regimes for hyper-parameter tuning and early stopping, namely (i) using the \textsc{En} COPA validation set or (ii) target language XCOPA validation set. Table~\ref{tab:ftsetups} lists all experimental setups.

\vspace{1.4mm}
\noindent \textbf{Neural Architecture.} 
As multiple--choice selection tasks differ in the number of choices (e.g., there are 2 possible answers in COPA, whereas there are 3 in SIQA), a classifier with a fixed number of classes is not a good fit for this scenario. We thus follow \citet{Sap:2019siqa} and couple the (pretrained) encoder with a feed-forward net which produces a single scalar score for each possible answer. The scores for individual choices are then concatenated and passed to the softmax function. 

As an input to the encoder, we couple each of the answer choices with the concatenation of the premise and the prompt. We feed this as a ``sentence pair'' input to MBERT and XLM-R, or as a single ``sentence'' to USE.\footnote{For MBERT and XLM-R, we insert the standard special tokens. For example, for the last example from Table~\ref{tab:causeeffect} and Choice 1, the input for MBERT would be as follows: ``\textit{[CLS] My eyes became red and puffy. What was the cause? [SEP] I was sobbing. [SEP]}''.} 

Let $c_i$ be the $i$-th answer choice of an instance of multiple-choice dataset (i.e., $i \in \{1, 2\}$ in XCOPA and $i \in \{1, 2, 3\}$ in SIQA) and let $\mathbf{x}_i \in \mathbb{R}^H$ (with $H$ as the vector size of the encoder)\footnote{For MBERT Base and XLM-R Base, $H = 768$; for XLM-R Large, $H=1,024$; for multilingual USE Large, $H = 512$.} be the encoding of its corresponding input consisting of the premise, prompt and the answer itself, as explained above.\footnote{For MBERT and XLM-R $\mathbf{x}_i$ is the transformed representation of the sequence start token. For USE, $\mathbf{x}_i$ is the average of contextualised vectors of all tokens.} The predicted score $\hat{y}_i$ for the answer $c_i$ is then obtained with the following feed-forward net: $\hat{y}_i = \mathbf{W}_o \tanh \left(\mathbf{W}_h \mathbf{x}_i + \mathbf{b}_h\right)$,
%
%{\footnotesize
%\begin{equation}
%    \hat{y}_i = \mathbf{W}_o \tanh \left(\mathbf{W}_h \mathbf{x}_i + \mathbf{b}_h\right)   
%\end{equation}}
%%
with $\mathbf{W}_h \in \mathbb{R}^{H \times H}$, $\mathbf{b}_h \in \mathbb{R}^{H}$ and $\mathbf{W}_o \in \mathbb{R}^{1 \times H}$ as parameters. We obtain the score $\hat{y}_i$ for each answer $c_i$ and concatenate them into a prediction vector to which we apply softmax normalisation: $\hat{\mathbf{y}} = \mathrm{softmax}([\hat{y}_1, \dots, \hat{y}_N])$, where $N$ is the number of answers in the multiple-choice classification dataset. We minimise the cross-entropy loss function via stochastic gradient descent.  

%% (IV, Removed, implementation details)
%% \footnote{We load \texttt{bert-base-multilingual-cased} (MBERT) and \texttt{xlm-roberta-base} (XLM-R) from HuggingFace Transformers. We precompute (static) USE vectors with the pretrained model from the TensorFlow Hub.}

\setlength{\tabcolsep}{7pt}
\begin{table}[t]
    \centering
    \def\arraystretch{1.0}
    {\footnotesize
    \begin{tabularx}{0.48\textwidth}{l YY YY}
    \toprule 
    & \multicolumn{2}{c}{Train dataset} & \multicolumn{2}{c}{Valid dataset} \\
    \cmidrule(lr){2-3} \cmidrule(lr){4-5}
    Setup & SIQA & COPA & \textsc{en} & target \\
    \cmidrule(lr){2-3} \cmidrule(lr){4-5}
    CO-ZS & & $\checkmark$ & $\checkmark$ & \\
    CO-TLV & & $\checkmark$ & & $\checkmark$ \\
    SI-ZS & $\checkmark$ & & $\checkmark$ & \\
    SI+CO-ZS & $\checkmark$ & $\checkmark$ & $\checkmark$ & \\
    SI+CO-TLV & $\checkmark$ & $\checkmark$ & & $\checkmark$ \\
    
    \bottomrule
    \end{tabularx}
}%
    \vspace{2mm}
    \caption{Different experimental setups for data sources. CO=COPA; SI=SIQA; ZS=Zero-Shot; TLV=Target Language Validation (Set).}
    \label{tab:ftsetups}
    \vspace{1mm}
\end{table}
 
\subsection{Results and Discussion}
\label{ssec:resdisc}

\setlength{\tabcolsep}{5pt}
\begin{table}[t]
    \centering
    {\footnotesize
    \begin{tabularx}{\linewidth}{c l c c c}
    \toprule
\textbf{Setup} & \textbf{Model} & All & {\scriptsize \shortstack{MBERT $\cap$ \\ XCOPA}} & {\scriptsize \shortstack{USE $\cap$ \\ XCOPA}} \\ \midrule
\multirow{3}{*}{CO-ZS} & XLM-R & 55.6 & 56.9 & 55.4 \\
& XLM-R-L & 52.4 & 52.5 & 52.1\\
& MBERT & 54.1 & 54.4 & 55.7 \\
& USE & 54.7 & 56.0 & 58.1 \\ \midrule
%%%%%
%%%%%
%%%%%
\multirow{3}{*}{CO-TLV} & XLM-R & 55.1 & 56.4 & 55.2\\
& XLM-R-L & 51.6 & 51.7 & 52.1 \\
& MBERT & 54.2 & 54.5 & 55.8 \\
& USE & 54.8 & 55.4 & 59.0 \\ \midrule
%%%%%
%%%%%
%%%%%
\multirow{3}{*}{SI-ZS} & XLM-R & 60.1 & 62.3 & 62.9 \\
& XLM-R-L & 68.4 & 72.1 & 72.9 \\
& MBERT & 54.7 & 55.6 & 56.4 \\
& USE & 55.0 & 56.4 & 60.1 \\ \midrule
%%%%%
%%%%%
%%%%%
\multirow{3}{*}{\shortstack{SI+CO-ZS}} & XLM-R & 59.0 & 60.7 & 61.9 \\
& XLM-R-L & 67.3 & 70.8 & 71.8 \\
& MBERT & 55.8 & 56.8 & 57.9 \\
& USE & 54.1 & 54.9 & 58.9 \\ \midrule
%%%%%
%%%%%
%%%%%
\multirow{3}{*}{\shortstack{SI+CO-TLV}} & XLM-R & 60.7 & 63.5 & 63.6 \\
& XLM-R-L & \textbf{69.1} & \textbf{72.8} & \textbf{74.6} \\
& MBERT & 54.4 & 54.8 & 54.2 \\
& USE & 54.3 & 55.2 & 59.1 \\
%%%%%
\bottomrule
    \end{tabularx}
}
    \vspace{2mm}
    \caption{Summary of XCOPA results. \textbf{All}: average over all 11 XCOPA languages; \textbf{MBERT\,$\cap$\,XCOPA}: average over 9 XCOPA languages (without \textsc{ht} and \textsc{qu}) included in MBERT and XLM-R pretraining; \textbf{USE\,$\cap$\,XCOPA}: average over 4 XCOPA languages (\textsc{it}, \textsc{th}, \textsc{tr}, and \textsc{zh}), included in the USE pretraining.}
    \label{tab:results}
    \vspace{1mm}
\end{table}

We first present the results for model transfer based on multilingual pretrained encoders. Table~\ref{tab:results} shows the aggregate accuracy of MBERT, XLM-Rs and USE over 11 XCOPA languages for each of the previously described training setups from Table~\ref{tab:ftsetups}. We first compare our cross-lingual average XCOPA results in the best setup with the English COPA performance of the monolingual English BERT (Base) reported by \newcite{Sap:2019siqa}, namely 63 accuracy in COPA-only fine-tuning (+7\%) and 80 after sequential SIQA\,+\,COPA fine-tuning (+17\%). This contributes to recent suspicions \cite{cao2020multilingual,Hu:2020arxiv} that massively multilingual pretrained transformers do not offer a completely satisfactory solution for language transfer.
\begin{figure*}[!t]
    \centering
    \includegraphics[width=0.99\linewidth]{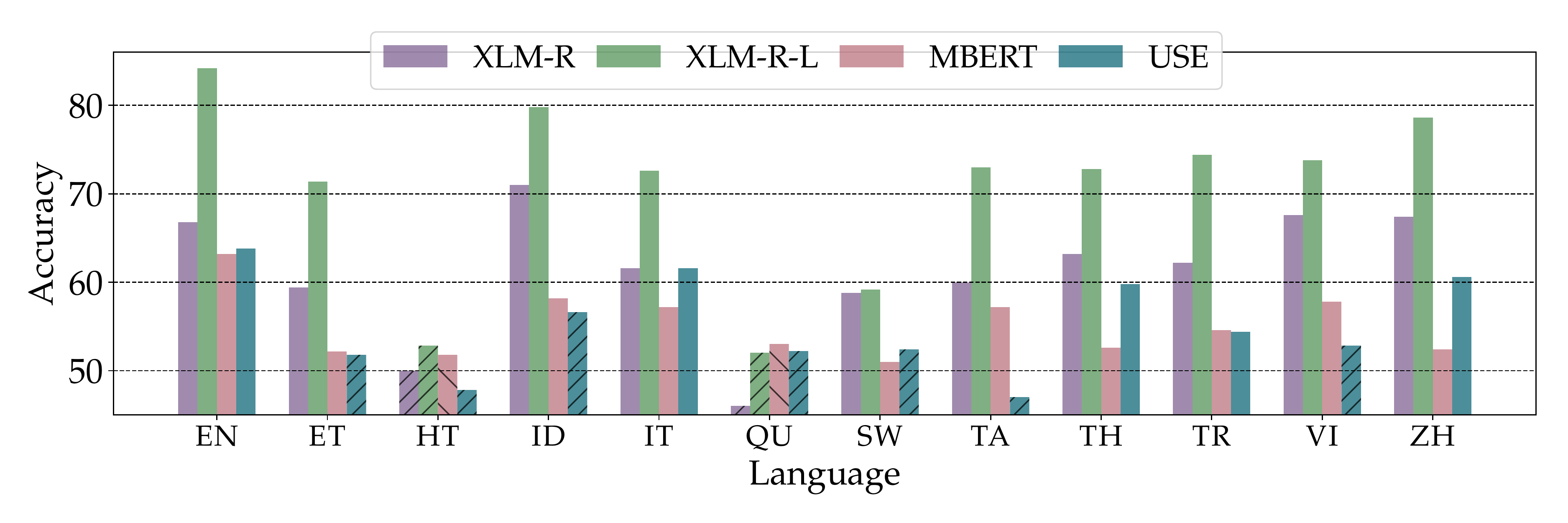}
    \vspace{-3.5mm}
    \caption{Per-language XCOPA results for XLM-R, MBERT, and USE in the SIQA\,+\,COPA-TLV setup. Striped bars correspond to language-model pairs where the language was not included in model pretraining.}
    \label{fig:lang_res}
    \vspace{-1.5mm}
\end{figure*}

\vspace{1.4mm}
\noindent \textbf{Multilingual Encoders.} XLM-R (both Base and Large) outperforms MBERT and USE in all setups, but the gains are pronounced only where the models were first fine-tuned on SIQA (SI-ZS, SI+CO-ZS, and SI+CO-TLV). USE outperforms MBERT often, which is especially surprising in the few-shot learning setup of COPA-only, as it contradicts the received wisdom that a larger amount of trainable parameters guarantees higher sample efficiency \citep{kaplan2020scaling}.
% Edo: I think the expectation is the opposite: the *larger* the number of encoder parameters, the *smaller* the number of parameters needed
%This might have been expected in the COPA-only setups (CO-ZS and CO-TLV) where the small COPA training set is insufficient to meaningfully fine-tune MBERT classifier parameters. 
%However, the finding that MBERT does not benefit more than USE from prior SIQA training is surprising and warrants further investigation. 
What is more, USE in some setups even surpasses MBERT for some of the languages (e.g., \textsc{id}, \textsc{ta}, \textsc{sw}) on which MBERT was pretrained and USE was not (cf.\,the scores in the MBERT\,$\cap$\,XCOPA column). We speculate that this is due to a combination of two effects: (1) the infamous ``curse of multilinguality'' \cite{conneau2019unsupervised} is much more pronounced for MBERT (which is pretrained on 104 languages) than for USE, pretrained on only 16 languages; and (2) there are subword-level similarities between XCOPA target languages and the 16 languages used in USE pretraining.
%
%% as well as the detailed per-language results in the Appendix
%

Unsurprisingly, XLM-R Large substantially outperforms its Base counterpart in all setups with SIQA training. Due to almost 3 times more parameters (355M vs.\,125M), XLM-R-L stores more language-specific information during pretraining. The large parameter space, however, also causes XLM-R-L to under-perform XLM-R in COPA-only setups (CO-ZS and CO-TLV), with the small COPA fine-tuning dataset.

\vspace{1.4mm}
\noindent \textbf{Data Sources.} 
Training models only on SIQA yields performance that is comparable (and for MBERT and USE often better) to performance we obtain with additional COPA training (setups SI\,+\,CO-ZS and SI\,+\,CO-TLV). While this is in part due to the limited size of the COPA training set, it confirms the assumption that SIQA and COPA are compatible tasks. We also note that only slight gains are achieved by hyper-parameter tuning on the target language validation set (TLV).  
 
% Overall, XLM-R is the best
% SIQA pretraining is very beneficial, suggesting that the small size of the COPA training set is particularly problematic

% Model selection using the target dev set useful for USE and XLM-R if pre-trained on SIQA. 

\vspace{1.4mm}
\noindent \textbf{Per-Language Performance.}
In Figure~\ref{fig:lang_res}, we report the language-specific performances for the best setup, SIQA\,+\,COPA-TLV, while we provide detailed results for all other setups in the Appendix. %~\ref{ss:resultsperlang}.
%Interestingly, the only XCOPA languages where MBERT performs better than XLM-R (and XLM-R performs worse than a random baseline) are \textsc{ht} and \textsc{qu}, the two out-of-sample languages.
As expected, all models fluctuate around random-level accuracy on the two out-of-sample languages, \textsc{ht} and \textsc{qu}.
%For all other languages, XLM-R outperforms MBERT.
Surprisingly, we also observe that for some languages (\textsc{id}, \textsc{vi}, \textsc{zh}) performance of transfer from English is slightly higher than the actual performance in English, without transfer. Moreover, the scores are often better for languages typologically distant from English than for closer ones (e.g., \textsc{th}, \textsc{vi}, \textsc{zh} vs. \textsc{it}). We speculate that this is due to the fact that languages such as \textsc{zh} and \textsc{th} are better represented in the pretrained models due to their unique scripts, which are shared with few other languages and therefore prevent lexical interferences.

\vspace{1.4mm}
\noindent \textbf{Cross-Lingual Transfer Methods.} Finally, in Table~\ref{tab:adv_res} (bottom) we compare the best setup for multilingual model transfer (XLM-R-L encoder with SIQA\,+\,COPA-TLV) with the method based on machine translation of the test set into English. In particular, for machine translation we consider both the massively multilingual (XLM-R-L) and the monolingual (RoBERTa Large, R-L) versions of the best encoder found in previous experiments.\footnote{As a caveat, note that these results are not perfectly comparable as Google Translate was trained on different (possibly more abundant) data, in addition to parallel texts.} No clear pattern emerges as the same model, XLM-R-L, is superior in 5 languages under the translation-based setup, and in 4 languages for the multilingual model transfer setup. A boost in accuracy, however, is especially evident for \textsc{sw}, with a gain of 13.8 points (+25\%). On the other hand, the translation-based setup largely outpaces multilingual model transfer when paired with a monolingual encoder, R-L. This baseline surpasses the others in all languages except for \textsc{th}. This brings us to the conclusion that it is not the cross-lingual transfer method in itself, but rather the avoidance of the `curse of multilinguality,' that makes translation-based cross-lingual transfer superior.

\subsection{Adversarial Variants}
\label{ssec:advvar}
After detecting an effective baseline, we `stress test' its robustness to prove that true causal reasoning is needed to solve our task. Recently, \citet{niven-kao-2019-probing} demonstrated that state-of-the-art pretrained encoders tend to rely on spurious correlations in natural language understanding datasets. Inspired by their work, we create two `adversarial' variants of XCOPA (and SIQA) where part of the input is masked. In the first variant (NoP), we hide premises; in the second (NoQ), we hide prompts. 

The results of the best configuration for multilingual model transfer are reported in Table~\ref{tab:adv_res} (top). Clearly, without premises the performance is not much better than chance (accuracy of 50.0). Without prompts (\textsc{cause} and \textsc{effect} for XCOPA), performance is higher but still lags behind the original variant with full inputs. Hence, we may conclude that solving this task requires causal reasoning over all components of the event dynamics.

\setlength{\tabcolsep}{6.5pt}
\begin{table*}[t]
    \centering
    {\footnotesize
    \begin{tabularx}{\textwidth}{l l l c c c c c c c c c c c}
    \toprule
\textbf{Setup} & \textbf{Variant} & \textbf{Model} & EN & ET & HT & ID & IT & SW & TA & TH & TR & VI & ZH \\ \midrule
MuMoTr & NoP & XLM-R-L & 63.0 & 56.2 & - & 62.4 & 56.6 & 56.0 & 60.2 & 60.6 & 58.8 & 60.2 & 63.2 \\
MuMoTr & NoQ & XLM-R-L & 76.2 & 63.8 & - & 73.4 & 72.8 & 59.0 & 64.4 & 70.2 & 69.2 & 69.8 & 70.8 \\

\hline
MuMoTr & Full & XLM-R-L & 84.2 & 71.4 & - & 79.8 & 72.6 & 59.2 & 73.0 & \textbf{72.8} & 74.4 & 73.8 & 78.6 \\
TrTe & Full & XLM-R-L & 84.2 & 76.8 & 70.0 & 79.6 & 74.6 & 74.0 & 74.0 & 71.4 & 72.8 & 77.8 & 78.4 \\
TrTe & Full & R-L & \textbf{88.4} & \textbf{81.0} & \textbf{73.8} & \textbf{82.2} & \textbf{77.8} & \textbf{74.2} & \textbf{79.6} & 71.4 & \textbf{79.6} & \textbf{81.0} & \textbf{86.0} \\

%%%%%
\bottomrule
    \end{tabularx}
}
    \caption{Detailed per-language XCOPA results of the best cross-lingual transfer setups (bottom). Performance with two adversarial variants of the dataset where premises and prompts are hidden, respectively (top).}
    \label{tab:adv_res}
    \vspace{-1em}
\end{table*}

\subsection{Adaptation to Unseen Languages}
\label{s:extension}
Even massively multilingual encoders like MBERT and XLM-R, pretrained on corpora of over 100 languages, cover only a fraction of the world's 7,000+ languages. In fact, the majority of the world languages suffer from data paucity \citep{kornai2013digital}: we thus explore several resource-lean approaches for extending encoders post-hoc to support transfer to languages not observed during pretraining, such as \textsc{qu} and \textsc{ht} in XCOPA.

%, i.e., languages the multilingual transformer was not exposed to during pretraining. 

\vspace{1.5mm}
\noindent \textbf{Adaptation Strategies.}
We use XLM-R, the best-performing among the ``base size'' multilingual encoders on XCOPA, and probe several strategies for adapting it to the two unseen XCOPA target languages. In all strategies, we simply continue training the XLM-R model via the masked language modeling (MLM) objective on different combinations of data \cite{Pfeiffer:2020emnlp}, in particular:

%% IV: MAD-X does the same -> mention it in the camera-ready
\vspace{1.4mm}
\noindent \textbf{1) T.} Sentences in the target language. We create the monolingual corpora for \textsc{ht} and \textsc{qu} by concatenating their respective Wikipedia dumps with their respective text from the JW300 corpus \cite{agic-vulic-2019-jw300}. In total, the training size is 5,710,426 tokens for \textsc{ht}, and 2,263,134 tokens for \textsc{qu}.
\vspace{1.3mm}
%%.(in our case of XCOPA transfer, these are Haitian Creole and Quechua).        

%\vspace{0.5em} 

%\noindent \textbf{Target Language Masked Language Modeling (XLMR-T).} This is the most extension approach in which we simply continue training the pre-trained XLM-R model via masked language modeling on a (reasonably small) target language corpus. We created the training corpora for HT and QU by concatenating their respective Wikipedias with their respective text from the JW300 corpus \cite{agic-vulic-2019-jw300}.\footnote{In total, the training size is 5,710,426 tokens for Hatian is and 2,263,134 tokens for Quechua. } 

%\vspace{0.5em} 

%%\noindent \textbf{Source and Target Language Masked Language Modeling (XLMR-S+T).} Here we perform additional MLM pretraining on the corpus which is the concatenation of the source language (English) and target language corpora (HT/QU). This could prevent (catastrophic) forgetting of the source language, which presumably may occur in T-MLM, and facilitate the downstream transfer. We create the English corpus of comparable size to HT and QU corpora by randomly sampling 200K sentences from English Wikipedia.
\noindent \textbf{2) S.} Sentences in English (\textsc{en}). This could prevent (catastrophic) forgetting of the source language while fine-tuning, which presumably may occur with T only. We create the English corpus of comparable size to \textsc{ht} and \textsc{qu} corpora by randomly sampling 200K sentences from \textsc{en} Wikipedia.

%\vspace{0.5em} 

\vspace{1.3mm}
\noindent \textbf{3) D.} A bilingual \textsc{en--ht} and \textsc{en--qu} dictionary. The dictionaries were extracted from PanLex \cite{kamholz-etal-2014-panlex}: we retain the 5k most reliable word translation pairs according to the available PanLex confidence scores. We create a synthetic corpus from the dictionary (termed \textit{D-corpus} henceforth) by concatenating each translation pair from the dictionary into a quasi-sentence. %In the first variant, we just perform additional MLM training only on the synthetic D-corpus.

%we perform additional MLM training on concatenated monolingual source and target language corpora plus the synthetic D-corpus.

%In the first model, XLM-D, we just perform additional MLM training only on the synthetic D-corpus.

%\vspace{1.3mm}
%\noindent \textbf{4) XLMR-S+T+D.} It is similar to XLMR-S+T, but we now run MLM on concatenated monolingual corpora plus the synthetic D-corpus.

\vspace{1.3mm}
\noindent \textbf{4) T-REP.} T data with all occurrences of target language terms from the 5K dictionary replaced with their English translations.

\vspace{1.5mm}

\noindent We select 5k target language sentences as the development corpus and use it for early stopping the MLM training (with perplexity as a metric).

\setlength{\tabcolsep}{7pt}
\begin{table}[t]
    \centering
    {\footnotesize
    \begin{tabularx}{\linewidth}{c l Y Y}
    \toprule
\textbf{Setup} & \textbf{Model} & \textsc{ht} & \textsc{qu} \\ \midrule
%%%%
\multirow{6}{*}{\rotatebox[origin=c]{90}{CO-ZS}} & XLM-R & 49.4 & 50.7 \\  \cdashline{2-4}
& ~~~~+T & \textbf{53.8} & 49.8 \\
& ~~~~+S+T & 52.8 & 54.0 \\ 
& ~~~~+D & 52.2 & 51.2 \\ 
& ~~~~+S+T+D & 53.6 & 52.0 \\
& ~~~~+T-REP & 49.6 & \textbf{55.0} \\ \midrule
%%%%%
\multirow{6}{*}{\rotatebox[origin=c]{90}{SI-ZS}} & XLM-R & 49.2 & 51.0\\  \cdashline{2-4}
& ~~~~+T & 56.2 & \textbf{57.9} \\
& ~~~~+S+T & 55.2 & 55.0 \\ 
& ~~~~+D & 55.4 & 57.4 \\ 
& ~~~~+S+T+D & 56.4 & 53.5 \\
& ~~~~+T-REP & \textbf{58.6} & 57.7 \\ \midrule
%%%%% 
\multirow{6}{*}{\rotatebox[origin=c]{90}{SI+CO-ZS}} & XLM-R & 51.4 & 51.2 \\  \cdashline{2-4}
& ~~~~+T & 57.8 & 54.0 \\
& ~~~~+S+T & 55.8 & 55.2  \\ 
& ~~~~+D & 57.8 & \textbf{57.9} \\ 
& ~~~~+S+T+D & 55.4 & 54.0 \\
& ~~~~+T-REP & \textbf{58.4} & 54.4 \\
%%%%%
\bottomrule
\end{tabularx}
}%
    \vspace{-1.5mm}
    \caption{XCOPA accuracy scores of different transfer variants that adapt to out-of-sample languages.}
    \label{tab:extension_results}
    \vspace{-1.5mm}
\end{table}

\vspace{1.6mm}
\noindent \textbf{Results and Discussion.} The performance of the five adaptation variants with XLM-R on \textsc{ht} and \textsc{qu} in the zero-shot XCOPA evaluation setups is reported in Table~\ref{tab:extension_results}. 
% (SI-ZS, CO-ZS, and SI+CO-ZS) 
% which has not been exposed to any HT/QU data.
When using sufficiently large fine-tuning datasets (SI-ZS and SI+CO-ZS setups) all adaptation methods yield substantial improvements over the XLM-R Base model.  The improvements are less consistent in the COPA-ZS setup. However, we attribute this to the limited size of the English COPA training set (which contains mere 400 instances) used for fine-tuning rather than to the ineffectiveness of the proposed adaptation strategies. A comparison between XLM-R+T and XLM-R+S+T suggests that additional MLM pretraining on a moderately sized target language corpus does not lead to catastrophic forgetting of the source language information. %%which would lead to deteriorated (X)COPA transfer. 

%% is comparable to the performance of the original XLM-R for some of the seen XCOPA languages.\footnote{See e.g., the performance for ET, SW, or TA in Figure \ref{fig:lang_res} and in the Appendix tables)} We find these results to be

The results of the light-weight post-hoc XLM-R adaptations for \textsc{ht} and \textsc{qu} are encouraging,\footnote{Note that the unseen languages, however, must rely on seen scripts (e.g., both \textsc{ht} and \textsc{qu} are written in Latin script).} 
%as they suggest that it is possible to improve downstream cross-lingual transfer performance even for out-of-sample languages unseen during massively multilingual pretraining.
as they bypass retraining the encoder from scratch 
%on multilingual corpora augmented with the corpus of the new/added language. 
while achieving downstream results almost comparable with seen languages.
Also, the results in Table~\ref{tab:extension_results} suggest that leveraging additional knowledge from a general bilingual dictionary can lead to further benefits: e.g., note the results of XLM-R+T-REP in SIQA-ZS and SIQA+COPA-ZS transfer setups, as well as XLM-R+D. Building on this proof-of-concept experiment, further adaptation strategies for zero-shot learning may be explored in the future, such as conditioning parameters on typological features \citep{ponti2019towards}.

\section{Related Work}
\label{s:related}
%%\textcolor{red}{Ivan}
\noindent \textbf{Evaluation of Commonsense Reasoning.} 
Besides COPA, another important early dataset that instigated computational modeling of commonsense reasoning is the Winograd Schema Challenge \cite[WSC;][]{Levesque:2012wsc,Morgenstern:2015aaai}. WSC consists in a pronoun coreference resolution task with paired instances, 
%which is trivial for humans, but difficult for machines. Due to the difficulty of creating such examples, the original WSC comprised 272 instances, but a similar and much larger 
and has been recently expanded into the WinoGrande dataset \cite{Sakaguchi:2020aaai} through crowd-sourcing.%, now spanning 44k paired instances. 

%% (IV: We said this before now)
%%Evaluation in COPA is framed as a binary classification task focused on \textit{causality}: it comprises 1,000 questions (500 for development and 500 for testing) asking which of two candidate sentences reflect a \textit{cause} or an \textit{effect} of a provided \textit{premise}.

%Recent advances in a range of NLP tasks driven by large pre-trained language models \cite{Wang:2019neurips,Ruder:2019naacl} has spurred further interest in this area. %as a way to probe their reasoning abilities.
%such as BERT \cite{Devlin:2019bert}, GPT \cite{radford2019language}, XLNet \cite{Yang:2019xlnet}, and XLM-R \cite{conneau2019unsupervised} have intensified further research and evaluation set creation for commonsense reasoning. The main incentive is to probe how much and what type of commonsense knowledge is captured by the large unsupervised models \cite[\textit{inter alia}]{Klein:2019acl,Bosselut:2020acl,Davison:2019emnlp,Forbes:2019cogsci,Zhou:2020aaai,Malaviya:2020aaai,Rogers:2020primer}. 
Several recent evaluation sets target particular \ignore{well-defined} aspects of commonsense, e.g., abductive reasoning \cite{Bhagavatula:2020iclr},\footnote{Abductive reasoning is inference to the most plausible explanation of incomplete observations \cite{Peirce:1960book}. 
%Sherlock Holmes uses this method of reasoning, while he falsely refers to it as ``deductive reasoning'' \cite{Carson:2009holmes}. %\textcolor{red}{IV: We can remove the part on Sherlock Holmes, I just had to mention it :)}
} intents and reactions to events \cite{Rashkin:2018acl}, social \cite{Sap:2019siqa} and physical \cite{Bisk:2020aaai} interactions, or visual commonsense \cite{Zellers:2019cvpr}. Others, e.g., CommonsenseQA \cite{Talmor:2019naacl}, SWAG \cite{Zellers:2018emnlp}, and HellaSWAG \cite{Zellers:2019acl} are cast as open-ended multiple-choice problems %where the system needs to choose the most sensible option.
where the most sensible option is chosen.
%(e.g., by selecting the most plausible answer in CommonsenseQA, or the most likely follow-up sentence given the context as in SWAG data). 
Another line of evaluation involves commonsense-enabled reading comprehension and question answering \cite{Ostermann:2018lrec,Zhang:2018arxiv,Huang:2019emnlp}. 

\vspace{1.6mm}
\noindent \textbf{Multilingual Evaluation of Natural Language Understanding.} While the above commonsense reasoning datasets are limited to English, several multilingual datasets for other natural language understanding tasks are available, e.g.,
%Creation of multilingual benchmarks helps advance computational modeling of natural language understanding across different languages. 
%A natural and frequent task for the evaluation of cross-lingual representations is translation at the level of words (i.e., bilingual lexicon induction) or sentences (i.e., machine translation) \cite{Ruder:2019jair,Ruder:2019tutorial}. Recent work has recognised the need to create multilingual data across additional and diverse tasks such as 
%These include 
lexical semantic similarity \cite[Multi-SimLex;][]{Vulic:2020multisimlex}, document classification \cite[MLDoc;][]{Schwenk:2018lrec}, sentiment analysis \cite{Barnes:2018acl}, and natural language inference \cite[XNLI;][]{Conneau:2018emnlp}. Other recent multilingual sets target the QA task based on reading comprehension: MLQA \cite{Lewis:2019arxiv} in 7 languages, XQuAD \cite{Artetxe:2019arxiv} in 10; and TyDiQA \cite{tydiqa} in 11 typologically diverse languages. Further, PAWS-X \cite{Yang:2019emnlp} evaluates paraphrase identification in 6 languages. A standard and pragmatic approach to multilingual dataset creation is translation from an existing (English) dataset, e.g., Multi-SimLex from the extended English SimLex-999 \cite{Hill:2015cl}, XNLI from MultiNLI \cite{Williams:2018naacl}, XQuAD from SQuAD \cite{Rajpurkar:2016emnlp}, and PAWS-X from PAWS \cite{Zhang:2019naacl}. TyDiQA, however, was built independently in each language.
Finally, a large number of tasks has been recently integrated into unified multilingual evaluation suites, XTREME \cite{Hu:2020arxiv} and XGLUE \cite{Liang:2020xglue}. %However, despite the abundant and steadily growing list of multilingual evaluation sets, a comprehensive multilingual benchmark for commonsense reasoning in particular is still missing. XCOPA aims to fill this gap, respecting typological diversity by covering a representative language set, and also including some truly low-resource languages.

\section{Conclusion and Future Work}
\label{s:conclusion}
We presented the Cross-lingual Choice of Plausible Alternatives (XCOPA), a multilingual evaluation benchmark for causal commonsense reasoning. All XCOPA instances are aligned across 11 languages, which enables cross-lingual comparisons. The language selection was informed by variety sampling, in order to maximise diversity in terms of typological features, geographical macro-area, and language family. This allows us to test the robustness of machine learning models for an array of rare phenomena displayed by the chosen languages.

We also ran a series of cross-lingual transfer experiments, evaluating state-of-the-art transfer methods based on multilingual pretraining and fine-tuning on English. We observed that, although these methods perform better than chance, they still lag significantly behind the monolingual setting where test data are translated into English, due to the `curse of multilinguality.' %In addition, the transfer seems not to depend that much on the distance from the source, but rather on the abundance of target language data in multilingual pretraining. 
In addition, we verified that spurious correlations are insufficient to solve this new task by creating `adversarial' variants of the dataset, where premises or prompts are masked, thus showing that robust causal reasoning is indeed required to solve XCOPA.
Finally, we investigated resource-lean adaptation of pretrained multilingual models to out-of-sample languages, exploiting only small monolingual corpora and/or bilingual dictionaries, reporting notable gains.

We hope that this new challenging evaluation set will foster further research in multilingual commonsense reasoning and cross-lingual transfer.

%% We found that the extension is successful insofar as it leads to downstream performances close to the in-sample languages.

%We hope this new challenging evaluation set will foster further research in multilingual commonsense reasoning and cross-lingual transfer.

\section*{Acknowledgements}
This work is supported by the ERC Consolidator Grant LEXICAL (no 648909). EMP, IV, and AK are also funded through the Google Faculty Research Award 2018 for Natural Language Processing. GG is supported by the Eliteprogramm of the Baden-W\"{u}rttemberg Stiftung (AGREE Grant). Heartfelt thanks to Fangyu Liu, Ulla Meeri Petti, and Yi Zhu for their invaluable help.

\bibliography{refs}
\bibliographystyle{acl_natbib}

% FOR FINAL VERSION
\clearpage
\appendix
\section{Detailed Translation Guidelines}
\label{ss:dataguide}
Translation of the English COPA validation and test set instances into each of the 11 languages was carried out by a single translator per language, meeting the following eligibility criteria: (i) a native speaker of the target language, (ii) fluent in English, (iii) with minimum undergraduate education level. Each translator was presented with translation guidelines and a spreadsheet accessible online, containing one English premise-hypothesis triple per line, followed by an empty line where target translations were entered. The task consisted in (a) identifying the correct alternative for the English premise and (b) translating the premise and both alternatives into the target language, preserving the causal relations present in the original (see \S\ref{s:qualitative} for discussion of ambiguous and problematic cases). Each translator worked independently (using any external resources, such as English-target language dictionaries, if needed) and completed the task in its entirety, producing 100 validation and 500 test instance translations, and a label for each. To ensure the output preserves the lexical, temporal, and causal relations present in the original triples, the guidelines instructed to:

\begin{enumerate}[label=\roman*., itemsep=0pt, topsep=0pt]
\item maintain the original correspondence relations between lexical items, i.e., if the same English word appeared both in the premise and in the alternative choices (Premise: \textit{The friends decided to share the \underline{hamburger}.}; A1: \textit{They cut the \underline{hamburger} in half.}; A2: \textit{They ordered fries with the \underline{hamburger}.}), it was mapped into the same target-language equivalent in all three translated sentences;
\item ensure that the original chronology and temporal extension of events is preserved through appropriate choice of verbal tense and aspect in the target language, e.g., maintaining the distinction between perfective and imperfective aspect (Premise: \textit{My eyes became red and puffy.} [\textsc{perf}], A1: \textit{I was sobbing.} [\textsc{imperf}], A2: \textit{I was laughing.} [\textsc{imperf}]; See also \S \ref{s:qualitative} and Appendix \ref{ss:gramtense} for discussion of the challenges posed by tenseless languages);

\item in case of English words with no exact translations in the target language or referring to concepts absent from the target language culture (e.g., \textit{peach}), the following solutions were to be adopted, in order of preference: (1) using a common loanword from another language, provided it is understood by the general population of target-language speakers; (2) using a periphrasis to describe the same concept (e.g., \textit{a juicy fruit}); (3) substituting the original concept with a similar one that is more familiar to the target language speaker community (e.g., \textit{santol}), provided that it can play a similar role in the causal relations captured by the original premise--prompt--answers triple.  
\end{enumerate}

\noindent
The translators were encouraged to split the workload into multiple sessions with breaks in between. Additionally, translators were encouraged to provide feedback, commenting on translation challenges and solutions, which we discuss in \S\ref{s:qualitative}.

\iffalse
% Data stats; Time
\begin{table}[t]
    \centering
    \begin{tabularx}{1.0\columnwidth}{l|XXXXXXXXXXX}
    \toprule
& \textsc{et} & \textsc{ht} & \textsc{id} & \textsc{it} & \textsc{sw} & \textsc{ta} & \textsc{th} & \textsc{tr} & \textsc{vi} & \textsc{zh}\\
\hline
Time & ? & 20 & 15 & ? & 28 & 20 & ? & ? & 17 & ?\\
%\textit{test} & 98.2 & 96.4 & 98.8 & 97.0 & 87.0 & 98.6 & 98.2 & 96.4 & 98.4 & 96.6\\
\bottomrule
    \end{tabularx}
    \caption{Time spent [hrs] on translation per language. \alert{ Add missing durations.}}
    \label{tab:time}
\end{table}
\fi

\begin{figure*}[t]
    \centering
    \includegraphics[width=0.9\textwidth]{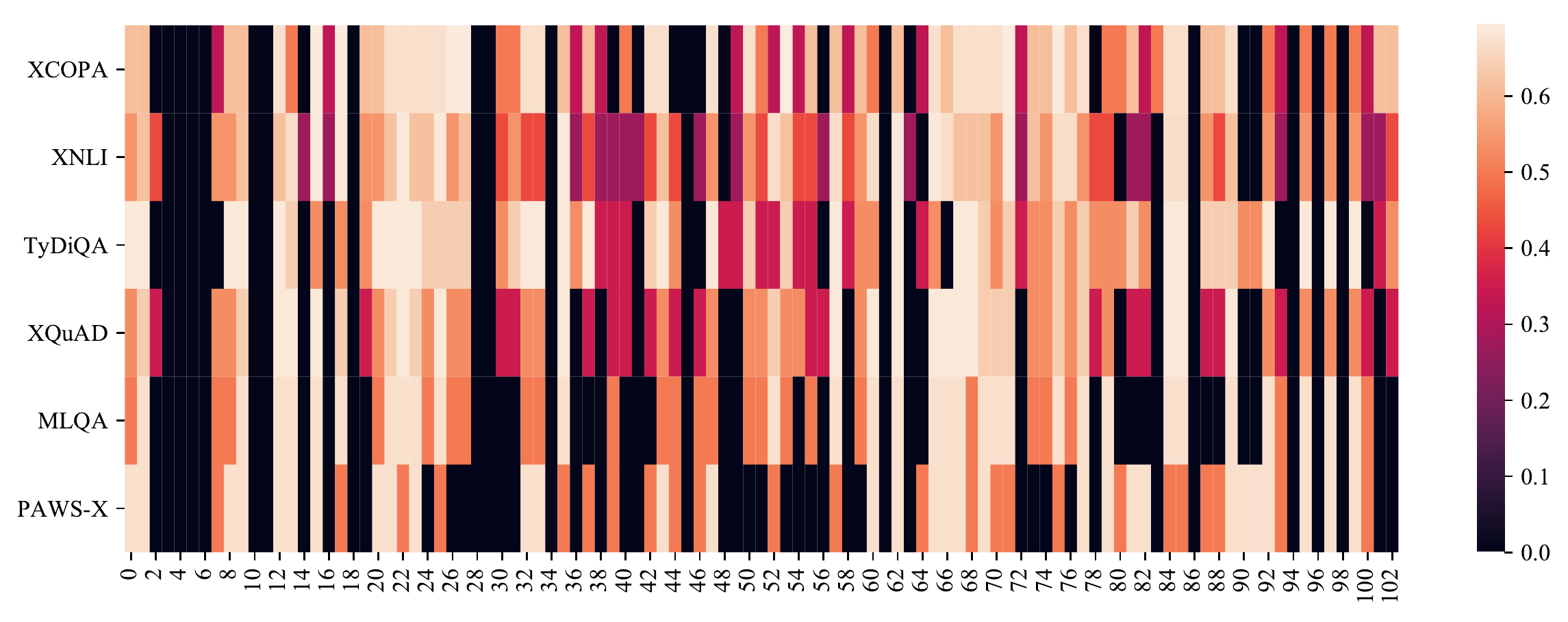}
    \vspace{-2.5mm}
    \caption{Heatmap of the entropy of the distributions of WALS features (x axis) in language samples from famous cross-lingual datasets outlined in \S\ref{s:related} (y axis).}
    \label{fig:entropiesbyfeat}
    \vspace{-1.5mm}
\end{figure*}
\setlength{\tabcolsep}{6.5pt}
\begin{table*}[t]
    \centering
    {\footnotesize
    \begin{tabularx}{\textwidth}{c l c c c c c c c c c c c c}
    \toprule
\textbf{Setup} & \textbf{Model} & EN & ET & HT & ID & IT & QU & SW & TA & TH & TR & VI & ZH \\ \midrule
\multirow{4}{*}{CO-ZS} & XLM-R & 57.6 & 59.8 & 49.4 & 58 & 56 & 50.7 & 57.2 & 56.6 & 52.8 & 56.2 & 58.5 & 56.6 \\
& XLM-R-L & 53 & 49.6 & 55.8 & 53 & 52.4 & 48 & 54 & 51.4 & 51.8 & 51 & 56 & 53 \\
& MBERT & 62 & 50.6 & 51.4 & 55 & 53.8 & 54.7 & 53.6 & 52 & 53.2 & 56.8 & 55.4 & 59  \\
& USE & 63 & 53.8 & 49.4 & 57.6 & 60 & 48.3 & 52.2 & 53 & 57.2 & 55 & 54.8 & 60.2 \\ \midrule
%%%%%
%%%%%
%%%%%
\multirow{4}{*}{CO-TLV} & XLM-R & 57.6 & 57.8 & 48.6 & 60.8 & 54.4 & 49.5 & 55.4 & 55.8 & 54.2 & 54.8 & 57.6 & 57.2 \\
& XLM-R-L & 53 & 49.4 & 47.8 & 51.4 & 53.6 & 54.2 & 50 & 47.8 & 53 & 50.6 & 58.2 & 51 \\
& MBERT & 62 & 52 & 52.6 & 58.2 & 55 & 52.7 & 53 & 52 & 52.4 & 53.8 & 52.6 & 61.8 \\
& USE & 63 & 49.4 & 49.6 & 57.6 & 62 & 54 & 50.8 & 53.6 & 58.6 & 56.2 & 51.4 & 59.2 \\ \midrule
%%%%%
%%%%%
%%%%%
\multirow{4}{*}{SI-ZS} & XLM-R & 68 & 59.4 & 49.2 & 67.2 & 63.6 & 51 & 57.6 & 58.8 & 61.6 & 60.4 & 65.8 & 66 \\
& XLM-R-L & 85 & 70.4 & 53.4 & 79.4 & 72.8 & 50.2 & 60.8 & 71 & 69.4 & 71.2 & 76 & 78.2 \\
& MBERT & 62.2 & 55.2 & 51.4 & 57 & 57 & 50.2 & 51 & 52.2 & 51 & 53.2 & 59.2 & 64.4 \\
& USE & 62.6 & 51.6 & 46.8 & 60.2 & 61.8 & 50.5 & 52.4 & 48.8 & 60.8 & 54.6 & 54.8 & 63 \\ \midrule
%%%%%
%%%%%
%%%%%
\multirow{4}{*}{\shortstack{SI+CO-ZS}} & XLM-R & 66.8 & 58 & 51.4 & 65 & 60.2 & 51.2 & 52 & 58.4 & 62 & 56.6 & 65.6 & 68.8 \\
& XLM-R-L & 84.2 & 68.8 & 52.8 & 79.8 & 72.4 & 50.7 & 59.4 & 68.2 & 67.2 & 71.2 & 73.8 & 76.2 \\
& MBERT & 63.2 & 52.2 & 54 & 59.4 & 57.2 & 48 & 56 & 54.6 & 51.2 & 57.4 & 58 & 65.6 \\
& USE & 63.8 & 51.2 & 48.4 & 57.6 & 61.8 & 52 & 51.8 & 47 & 58 & 55.6 & 51 & 60.2 \\ \midrule
%%%%%
%%%%%
%%%%%
\multirow{4}{*}{\shortstack{SI+CO-TLV}} & XLM-R & 66.8 & 59.4 & 50 & 71 & 61.6 & 46 & 58.8 & 60 & 63.2 & 62.2 & 67.6 & 67.4 \\
& XLM-R-L & 84.2 & 71.4 & 52.8 & 79.8 & 72.6 & 52 & 59.2 & 73 & 72.8 & 74.4 & 73.8 & 78.6 \\
& MBERT & 63.2 & 52.2 & 51.8 & 58.2 & 57.2 & 53 & 51 & 57.2 & 52.6 & 54.6 & 57.8 & 52.4 \\
& USE & 63.8 & 51.8 & 47.8 & 56.6 & 61.6 & 52.2 & 52.4 & 47 & 59.8 & 54.4 & 52.8 & 60.6 \\
%%%%%
\bottomrule
    \end{tabularx}
}
    \caption{Detailed per-language XCOPA results for multilingual model transfer. None of the encoders was exposed to \textsc{ht} and \textsc{qu} during pretraining. USE was exposed only to \textsc{it}, \textsc{th}, \textsc{tr}, and \textsc{zh}.}
    \label{tab:lang_res}
    \vspace{-1mm}
\end{table*}
\setlength{\tabcolsep}{5pt}
\begin{table*}[!t]
\def\arraystretch{0.93}
\centering
{\footnotesize
\begin{tabularx}{\textwidth}{l c l l X}
\toprule
{\bf Name} & {\bf Languages} & \textbf{Vocab} & {\bf Params} & {\bf URL} \\ \midrule
mBERT & 104 & 119K & 125M & {\url{https://huggingface.co/bert-base-multilingual-cased}} \\
XLM-R & 100 & 250K & 125M & {\url{https://huggingface.co/xlm-roberta-base}} \\
XLM-R-L & 100 & 250K & 355M & {\url{https://huggingface.co/xlm-roberta-large}} \\
\bottomrule
\end{tabularx}
}
\vspace{-1.5mm}
\caption{Pretrained transformers used in our study.}
\label{tbl:models}
\end{table*}

\section{Why is Grammatical Tense Problematic for XCOPA?}
\label{ss:gramtense}
The scenarios included in COPA refer to events that took place in the past and are formulated in what can be described as a narrative register: one of the sources from which question topics were drawn was a corpus of personal stories published online \cite{gordon2009identifying}. This is grammatically rendered exclusively by means of past simple (preterite) or past continuous (imperfect) verb forms. Temporal anteriority of a hypothesis sentence with respect to the premise is not grammatically marked (e.g., with a past perfect verb form) and can only be deduced based on the prompt (``\textit{What was the \textsc{cause} of this?}''). The preterite-imperfect contrast used in English to distinguish background states (imperfective) from the main event (perfective) (e.g., \textit{I was expecting company.} \textsc{imperf} vs. \textit{I tidied up my house.} \textsc{perf}) is not universally applicable and different languages employ different discourse grounding strategies \cite{hopper1979aspect}, which has interesting implications for the multilingual extension of COPA to XCOPA.

In the languages with grammatical tense, different strategies are employed to capture the perfective--imperfective distinction, which is prominent in COPA. For example, in Haitian Creole, the simple past marker \textit{te} is used to indicate a bounded event in the past, while the continuous aspect is signaled with an \textit{ap} marker. Italian additionally distinguishes between two perfective past tenses, expressed by means of a simple and compound past (\textit{vidi} - \textit{ho visto}, `I saw'). The opposition is between completed actions whose effects are detached from the present and those with persisting effects on the present. Both contrast with the imperfect, which emphasises the event's extension or repetition in time. Given that the opposition is a matter of the speaker's perspective on events and partially based on deixis (remote versus proximate past), the translator opted for the most natural choice given a specific context/situation.

\section{Hyper-Parameter Search}
For MBERT and XLM-R we searched the following hyper-parameter grid in both SIQA and COPA training: learning rate $\in \{5\cdot10^{-6}, 10^{-5}, 3\cdot 10^{-5}\}$, dropout rate (applied to the output layer of the transformer and the hidden layer of the feed-forward scoring net) $\in \{0, 0.1\}$, and batch size $\in \{4, 8\}$. For USE, we searched over different values for the learning rate, $\{10^{-3},10^{-4},10^{-5}\}$. We evaluate the performance on the respective development set every $500$ updates for SIQA and every $10$ updates for COPA and stop the training if there is no improvement after $10$ consecutive evaluations. In all setups, we optimise the parameters with the Adam algorithm \cite{kingma2015adam} with $\epsilon = 10^{-8}$, no weight decay, and no warm-up. We clip the norms of gradients in single updates to $1$.  

\section{Full Results (Per Language)} 
\label{ss:resultsperlang}
Table~\ref{tab:lang_res} contains the detailed per-language results of multilingual model transfer for all XCOPA languages and all five of our evaluation setups (CO-ZS, CO-TLV, SI-ZS, SI+CO-ZS, SI+CO-TLV).

%\twocolumn
\section{Code and Dependencies}

Our code is built on top of the HuggingFace Transformers framework\footnote{\url{github.com/huggingface/transformers}} and USE\footnote{\url{tfhub.dev/google/universal-sentence-encoder-multilingual-large/3}} and is available at \url{github.com/cambridgeltl/xcopa}. Table \ref{tbl:models} details the pretrained models that we exploited in this work. Besides these, our code relies only on standard Python's scientific computing libraries (e.g., \texttt{numpy}).

%\onecolumn

\end{document}